\documentclass{article}
\usepackage{spconf,amsmath,graphicx, pifont,url}

\newcommand{\cmark}{\ding{51}}%
\newcommand{\xmark}{\ding{55}}%
\title{PVDeConv: Point-Voxel DeConvolution for
Autoencoding CAD Construction in 3D}
\name{
Kseniya Cherenkova$^{\star \dagger}$
    \qquad
Djamila Aouada$^{\star}$
    \qquad
Gleb Gusev$^{\dagger}$
}
\address{$^{\star}$ SnT, University of Luxembourg
    \qquad
    $^{\dagger}$ Artec3D}
\begin{document}
%
\maketitle
\begin{abstract}
We propose a Point-Voxel DeConvolution (\textit{PVDeConv}) module for 3D data autoencoder. To demonstrate its efficiency  we learn to synthesize high-resolution point clouds of 10k points that densely describe the underlying geometry of Computer Aided Design (CAD) models. Scanning artifacts, such as protrusions, missing parts, smoothed edges and holes, inevitably appear in real 3D scans of fabricated CAD objects. Learning the original CAD model construction from a 3D scan requires a ground truth to be available together with the corresponding 3D scan of an object. To solve the gap, we introduce a new dedicated dataset, the CC3D, containing 50k+ pairs of CAD models and their corresponding 3D meshes. This dataset is used to learn a convolutional autoencoder for point clouds sampled from the pairs of 3D scans - CAD models. The challenges of this new dataset are demonstrated in  comparison with other generative point cloud sampling models trained on ShapeNet. The CC3D autoencoder is efficient with respect to memory consumption and training time as compared to state-of-the-art models for 3D data generation.
\end{abstract}
\section{Introduction}
\label{sec:intro}
Recently, deep learning (DL) for 3D data analysis has seen a boost in  successful and competitive solutions for segmentation, detection and classification {\cite{Guo2019DeepLF}, and real-life applications, such as self-driving, robotics, medicine, and augmented reality. In industrial manufacturing, 3D scanning of fabricated parts is an essential step of product quality control, when the 3D scans of real objects are compared to the original Computer Aided Design (CAD) models. 
While most consumer solutions for 3D scanning are good enough for capturing the general shape of an object, artifacts can be introduced in the parts of the object that are physically inaccessible for 3D scanning, resulting in the loss of sharp features and fine details. 
\begin{figure}
    \centering
    \includegraphics[width=\linewidth]{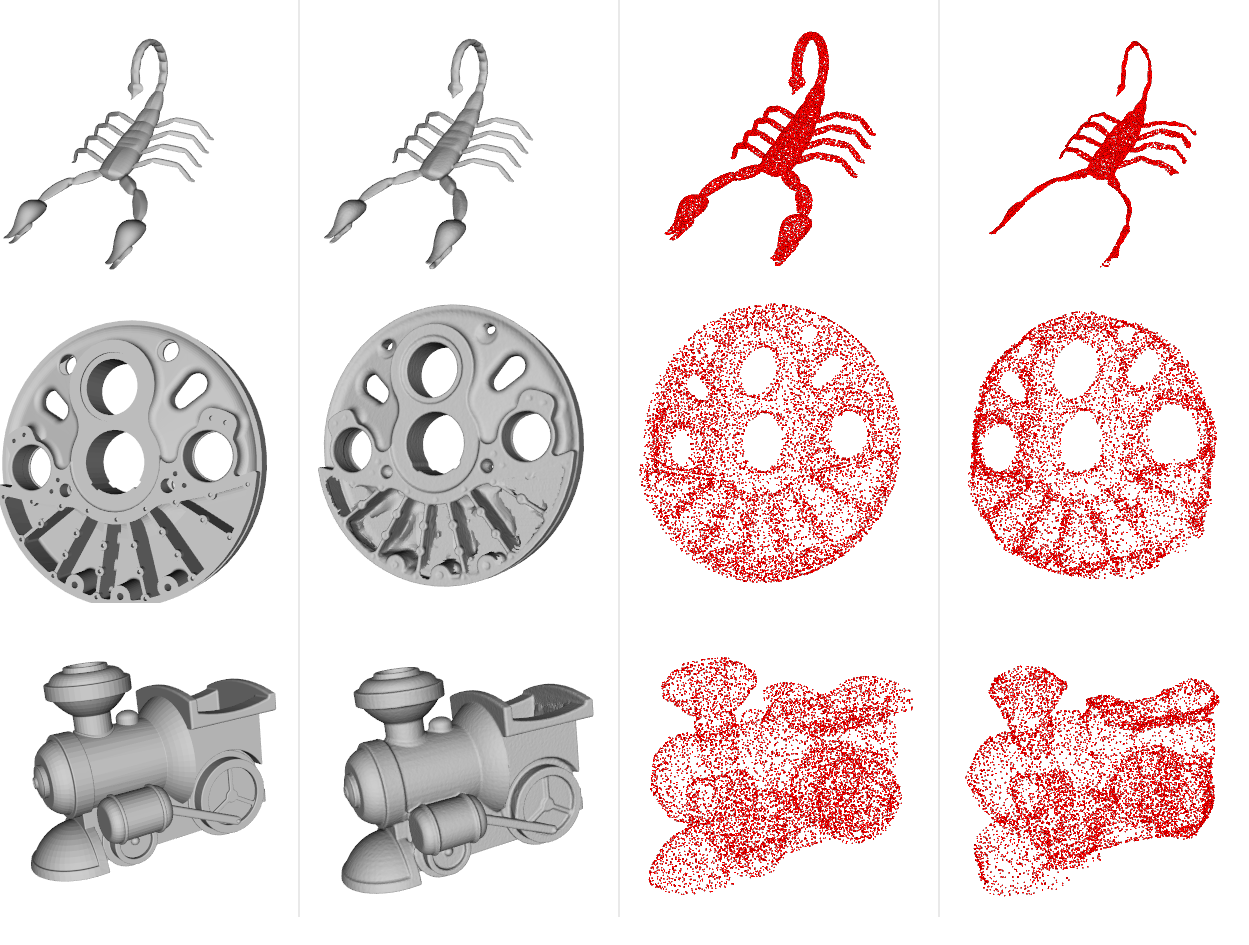}
    \vspace{-2em}
    \caption{
        Examples of CC3D data: From left to right, CAD models, corresponding 3D scans, 10k input point clouds and results of the proposed autoencoder.
    }
    \label{fig:cc3ddata}
    \vspace{-1em}
\end{figure}

This paper focuses on recovering scanning artifacts in an autoencoding data-driven manner. In addition to presenting a new point cloud autoencoder, we introduce a new 3D dataset, referred to as \emph{CC3D}, which stands for \emph{CAD Construction in 3D}. We further provide an analysis focused on real 3D scanned data, keeping in mind real-world constraints, i.e., variability, complexity, artifacts, memory and speed requirements. The first two columns in Fig.~\ref{fig:cc3ddata} give some examples from CC3D data; the CAD model and its 3D scanned version in triangular mesh format.
While the most recent existing solutions~\cite{atlasnet, topnet, ain, quadric} on 3D data autoencoders mostly focus on low-resolution data configuration (approximately 2500 points), we see it more beneficial for real data to experiment in higher-dimension. It is what brings the important 3D object details into the big data learning perspective.

Several publicly available datasets related to CAD modelling, such as ModelNet~\cite{modelnet}, ShapeNet~\cite{shapenet}, and ABC~{\cite{ABC}, have been released in the last years. The summary of the features they offer can be found in Table~\ref{table:datasets}. These datasets have boosted the research on deep learning on 3D point clouds mainly. 

Similarly, our CC3D dataset should support research efforts in addressing real-world challenges. Indeed, this dataset provides various 3D scanned objects, with their ground-truth CAD models. The models collected in CC3D dataset are not restricted to any object's category and/or complexity. 3D scans offer challenging cases of missing data, smoothed geometry and fusion artefacts in the form of varying protrusions and swept holes. Moreover, the resolution of 3D scans is typically high with more than 100k faces in the mesh.

In summary, the contributions of this paper include:
(1) A 3D dataset, CC3D, a collection of 50k+ aligned pairs of meshes, a CAD model and its virtually 3D scanned counterpart with corresponding scanning artifacts;
(2) A CC3D autoencoder architecture on 10k point clouds learned from CC3D data;
(3) A Point-Voxel DeConvolution (\emph{PVDeConv}) block for the decoder part of our model, combining point features on coarse and fine levels of the data.

The remainder of the paper is organized as follows: Section~\ref{sec:related} reviews relevant state-of-the-art works in 3D data autoencoding. In Section~\ref{sec:background} we give a brief overview of the core components our work is built upon. Section~\ref{sec:majhead} describes the main contributions of this paper in more details. In Section~\ref{sec:experiments} the results and comparison with related methods are presented. Section~\ref{sec:conclusions} gives the conclusions.

\section{Related Work}
\label{sec:related}
The choice of DL architecture and 3D data representation is usually defined by existing practices and available datasets for learning~\cite{Ahmed2018DeepLA}. Voxel-based representations have pioneered 3D data analysis, applying 3D Convolution Neural Network (CNN) directly on a regular voxel grid~\cite{VoxNet}. Despite the improved models in terms of memory consumption, e.g., ~\cite{OctNet}}, their inability to resolve fine object details remains the main limiting factor in practical use. 

Other works introduce convolutions directly on graph structures, e.g., ~\cite{COMA:ECCV18}. They attempt to generalize DL models to non-Euclidean domains such as graphs and manifolds~\cite{GDL}, and offer the analogs of pooling/unpooling operations as well~\cite{meshcnn}. However, they are not applicable for learning on real unconstrained data as they require either meshes to be registered to a common template, or inefficiently deal with meshes of up to several thousand faces, or are specific to segmentation or classification tasks only.

Recent advances in developing efficient architectures for 3D data analysis are mainly related to point cloud based methods~\cite{pointnet++, dgcnn}. Decoders {\cite{FoldingNet, atlasnet, 3dcapsule, topnet, pcn}} have made point clouds a highly promising representation for 3D object generation and completion using neural networks. Successful works in generative adversarial network (GAN)  (e.g.,{\cite{pointcloudgan}), show the applicability of different GAN models operating on the raw point clouds.

In this paper, we comply with the common autoencoder approach, i.e., we use a point cloud encoder to embed the point cloud input, and design a decoder to generate a complete point cloud from the embedding of the encoder. 
\section{BACKGROUND and MOTIVATION}
\label{sec:background}
We herein present the fundamental building blocks that comprise the core of this paper, namely, the point cloud, metric on it, and the DL backbone. All together, these elements make the CC3D autoencoder perform efficiently on high-resolution 3D data.

A point cloud $S$ can be represented as $S =\{(p_{k}, f_{k})\}$, where each $p_{k}$ is the 3D coordinates of the $k^{th}$ input point, and $f_{k}$ is the feature corresponding to it, and the size of $f_{k}$ defines the dimensionality of the points feature space. Note that while it is straightforward to include auxiliary information (such as points' normals) to our architecture, in this paper we exclusively employ the $xyz$ coordinates of $p_{k}$ as the input data. 

We base on Point-Voxel Convolution (PVConv), a memory efficient architecture for learning on 3D point cloud presented in {\cite{pvcnn}}. To the best of our knowledge, it is the first development of autoencoder based on PVCNN as the encoder. Briefly, PVConv  combines the fine-grained feature transformation on points with the coarse-grained neighboring feature aggregation in the voxel space of point cloud. Three basic operations are performed in the coarse branch, namely, voxelization, followed by voxel-based 3D convolution, and the devoxelization. The point-based branch aggregates the features for each individual with multilayer perceptron (MLP), providing high resolution details. The features from both branches are aggregated into a hidden feature representation.

The formulation of convolution in both voxel-based and point-based cases is the following:
\begin{equation}
    y_{k} = \sum_{x_{i} \in N(x_{k})} K(x_{k}, x_{i}) \times F(x_{i}),
    \label{equ:pvconv}
\end{equation}
where for each center $x_{k}$, and its  neighborhood $N(x_{k})$, the neighboring features $F(x_{i})$ are convolved with the kernel $K(x_{k}, x_{i})$.
The choice of PVCNN is due to its efficiency in training on high-resolution 3D data. Indeed, it makes it a good candidate for working with real-life data. As it is stated in~\cite{pvcnn}, PVConv combines advantages of point-based methods, reducing memory consumption, and voxel-based, improving the data locality and regularity. 

For the loss function, Chamfer distance~\cite{Fan:Chamfer} is used to measure the quality of the autoencoder. It is a differentiable metric, invariant to permutation of points in both ground-truth and target point clouds, $S_{G}$ and $S$, respectively. It is defined as follows:
\begin{equation}
    d_{CD}(S, S_{G}) = \sum_{x \in S} \min_{y \in S_{G}}{\| x - y \|}^2 + \sum_{y \in S_{G}} \min_{x \in S} {\| x - y \|}^2.
    \label{equ:chamfer}
\end{equation}
As it follows from its definition, no correspondence or equal number of points in $S$ and $S_{G}$ is required for the computation of $d_{CD}$, making it possible to work within different resolutions for the encoder and decoder.

\section{Proposed AUTOENCODING of 3D Scans to CAD Models}
\label{sec:majhead}
This paper studies the problem of 3D point cloud autoencoding in a deep learning setup, and in particular, the choice of the architecture of a 3D point cloud decoder for efficient reconstruction of point clouds sampled from corresponding pairs of 3D scans and CAD models.
\subsection{CC3D dataset}
\label{ssec:subhead}
The CC3D dataset of 3D CAD models was collected from a free online service for sharing CAD designs~\cite{3dcontentcentral}. In total, the collected dataset contains 50k+ models in STEP format, unrestricted to any category, with varying complexity from simple to highly detailed designs (see examples in Fig.~\ref{fig:cc3ddata}). These CAD models are converted to meshes, and each mesh was virtually scanned using proprietary 3D scanning pipeline developed by Artec3D~\cite{Artec3D}. The typical size of the resulting scans is in the order of 100K points and faces, while the meshes converted from CAD models are usually more than an order of magnitude lighter.
\begin{table}[!b]
\centering
\vspace{-2em}
\begin{tabular}{llllllll}
 Dataset & \#Models & \rotatebox{90}{CAD} & \rotatebox{90}{Curves} & \rotatebox{90}{Patches} & \rotatebox{90}{Semantics} & \rotatebox{90}{Categories} & \rotatebox{90}{3D scan}
 \\ 
 \hline
 CC3D (ours) & 50k+ & \cmark & \xmark & \xmark & \xmark & \xmark & \cmark \\
 ABC~\cite{ABC} & 1M+ & \cmark & \cmark & \cmark & \xmark &\xmark & \xmark\\
 ShapeNet~\cite{shapenet} & 3M+ & \xmark & \xmark & \xmark & \cmark & \cmark & \xmark\\
ShapeNetCore~\cite{shapenet} & 51k+ & \xmark & \xmark & \xmark & \cmark & \cmark & \xmark\\
ShapeNetSem~\cite{shapenet} & 12k & \xmark & \xmark & \xmark & \cmark & \cmark & \xmark\\
ModelNet~\cite{modelnet} & 151k+ & \xmark & \xmark & \xmark & \cmark & \cmark & \xmark\\
\end{tabular}
\caption{\label{table:datasets}  Summary of datasets with CAD-like data. Note that only ABC and CC3D offer CAD models in b-rep (boundary representation) format in addition to triangular meshes.}
\end{table}

In order to illustrate the uniqueness of our dataset, Table~\ref{table:datasets} summarizes the available CAD-like datasets and semantic information they provide. Unlike ShapeNet~\cite{shapenet} and ModelNet~\cite{modelnet}, the CC3D dataset is a collection of 3D objects unrestricted to any category, with the complexity varying from very basic to highly detailed models. One of the most recent datasets, the ABC dataset~\cite{ABC} would have been a valuable collection due to its size for our task if it had contained 3D scanned models alongside with ground-truth CAD objects.
The availability of CAD-3D scan pairings, the high-resolution of meshes and variability of the models make the CC3D dataset stand out among other alternatives. The CC3D dataset will be shared with the research community.
\subsection{CC3D Autoencoder}
\label{sssec:subsubhead}
Our decoder is a modified version of PVCNN, where we cut the final classification/segmentation layer. The proposed PVDeConv structure is depicted in Fig.~\ref{fig:pvdeconv}. 
The fine point-based branch is implemented as shared transposed MLP, allowing to maintain the same number of points throughout the autoencoder, while the coarse branch allows the features to be aggregated at different voxel grid resolutions, thus modelling the neighborhood information at different scales. 

The PVDeConv block consists of 3D volumetric deconvolutions to aggregate the features, dropout, the batch normalization and the nonlinear activation function after each 3D deconvolution. Features from both branches are fused at the final level and MLP to produce the output points.

\begin{figure}
    \centering
    \includegraphics[width=\linewidth]{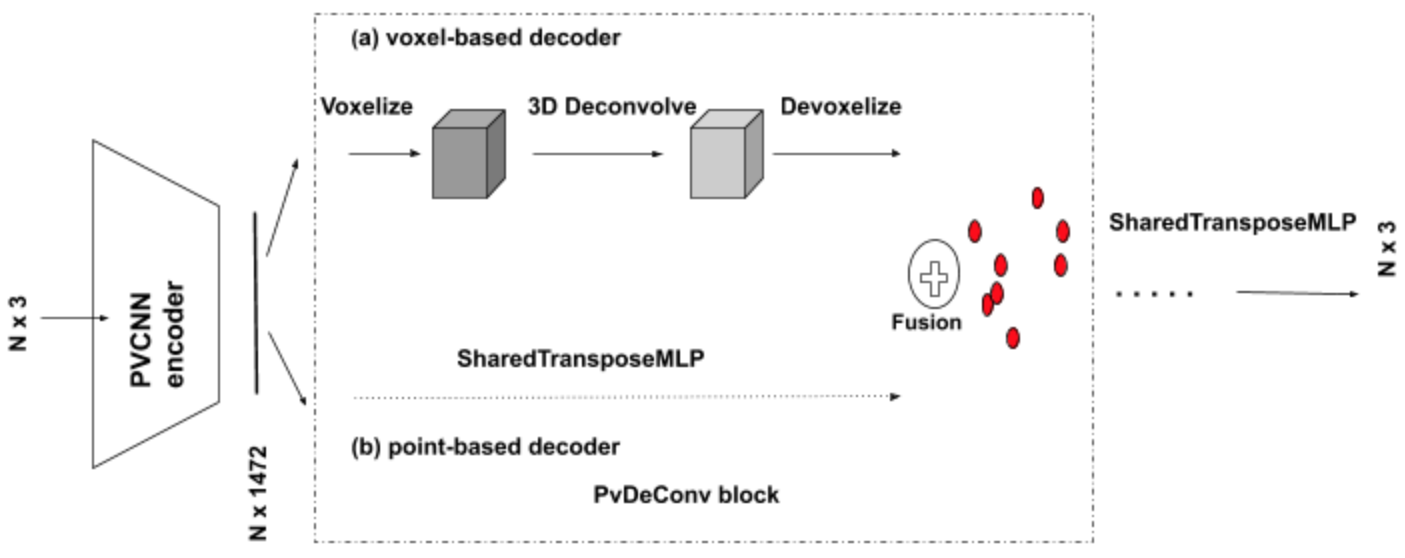}
    \vspace{-2em}
    \caption{Overview of CC3D autoencoder architecture and PVDeConv module. The features from coarse voxel-based and fine point-based branches are fused to be unwrapped to the output point cloud. 
    }
    \label{fig:pvdeconv}
    \vspace{-1em}
\end{figure}

The transposed 3D convolution operator, used in PVDeConv,  multiplies each input value element-wise by a learnable kernel, and sums over the outputs from all input feature channels. This operation can be seen as the gradient of 3D convolution, although it is not an actual deconvolution operation. 

\section{EXPERIMENTS}
\label{sec:experiments}
We evaluate the proposed autoencoder by training first on our CC3D dataset, and then on the ShapeNetCore~\cite{shapenet} dataset.
\subsection{Training on CC3D}
\quad\quad\textbf{\textit{Dataset.}} CC3D dataset is randomly split into three non-intersecting folds: 80\% for training, 10\% for testing and 10\% for validation.  Ground-truth point clouds are generated by uniformly sampling $N = 10$k points on the CAD models surfaces, while the input point clouds are sampled in the same manner from corresponding 3D scans of the models. The data is normalized to (0, 1).

\textbf{\textit{Implementation Details.}} The encoder follows the structure in {\cite{pvcnn}}, the coarse blocks are ((64, 1, 32), (64, 2, 16), (128, 1, 16), 1024), where triplets describe voxel-based convolutional PVConv block in terms of number of channels, number of blocks, and voxel resolution. The last number describes the resulting embedding size for the coarse part, and being combined with shared MLP cloud blocks = (256, 128), gives the feature embedding size of 1472. The decoder coarse blocks are ((128, 1, 16), (64, 2, 16), (64, 1, 32), 128), where the triplets are PVDeConv concatenated with decoder point-based fine blocks of size (256, 128). 

\textbf{\textit{Training setup.}} The autoencoder is trained with Chamfer loss for 50 epochs on two Quadro P6000 with batch size 80 in data parallel mode. The overall training takes approximately 15 hours. The best model is chosen based on the validation set.

\textbf{\textit{Evaluation.}} The qualitative results of our autoencoder on the CC3D data are presented in Fig.~\ref{fig:cc3dres}. We notice that the fine details are captured in these challenging cases.
\begin{figure}
    \centering
    \vspace{-2em}
    \includegraphics[width=\linewidth]{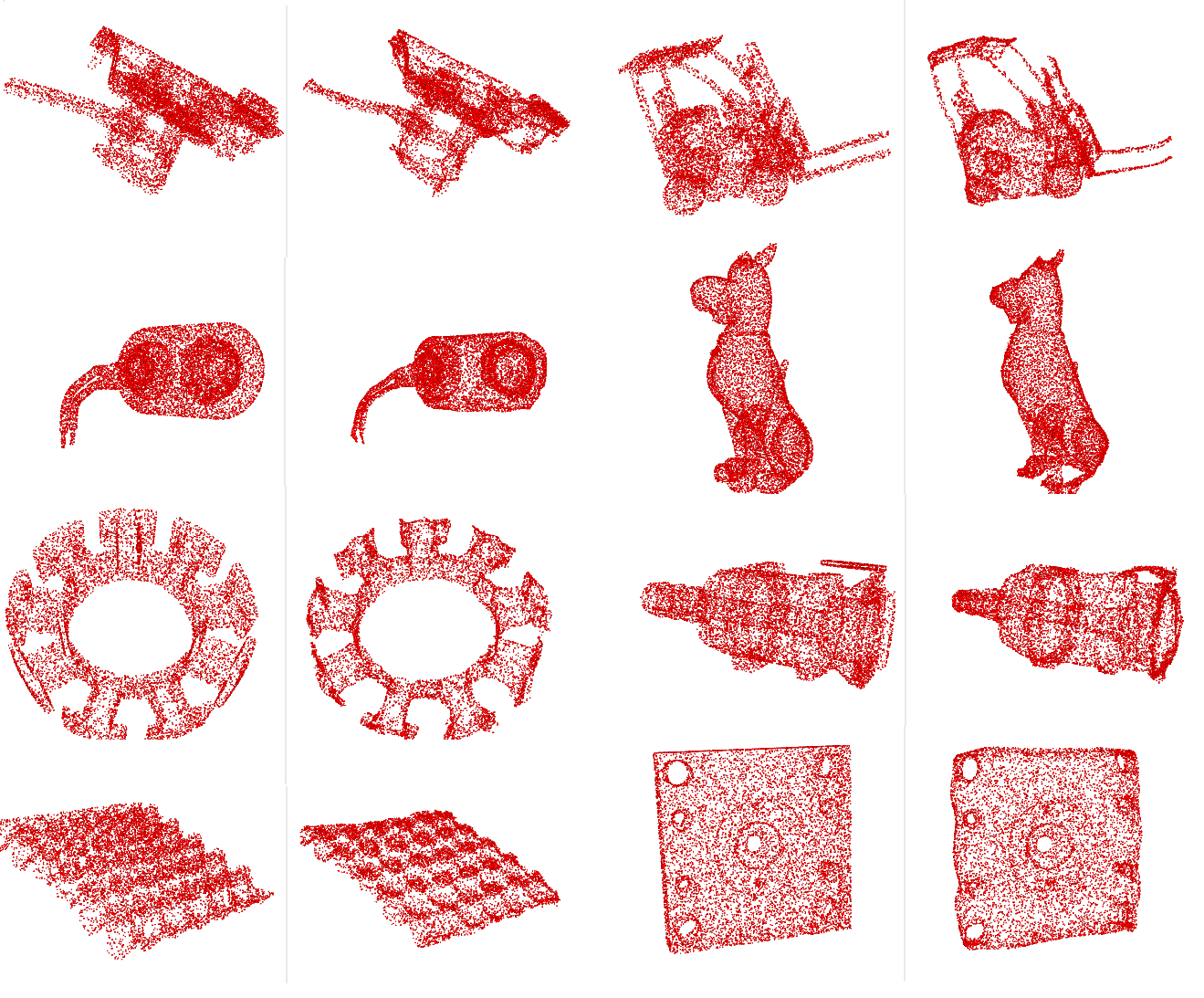}
    \caption{Results of our autoencoder on CC3D data with 10k points for input and output. The left of each pair of results is the input point cloud of 10k, the right is the autoencoder reconstruction of 10k points.
    }
    \label{fig:cc3dres}
    \vspace{-1em}
\end{figure}
%
\begin{table}[h]
    \centering
    \begin{tabular}{c|c}
    Method & Chamfer distance, $10^{-3}$ 
    \\ \hline
        AtlasNet {\cite{atlasnet}} & 1.769 \\
        FoldingNet {\cite{FoldingNet}} & 1.648 \\
        PCN {\cite{pcn}} & 1.472\\
        TopNet {\cite{topnet}} & 0.972 \\
        \textbf{Ours} & \textbf{0.804}
    \end{tabular}
    \caption{CC3D autoencoder results on ShapeNetCore dataset: comparison against previous works ($N=2.5$k).}
    \label{tab:chamferShapeNet}
    \vspace{-1em}
\end{table}
\subsection{Training on ShapeNetCore}
To demonstrate the competitive performance of our CC3D autoencoder, we train it on the ShapeNetCore dataset following the train/test/val split of~\cite{topnet}, with the number of sampled point $N=2500$ for a fair comparison. Since we do not have scanned models for the ShapeNet data, we add a 3\% Gaussian noise to each point's location. The rest of the training setup is replicated from the CC3D configuration. The final metric is the mean Chamfer distance averaged per model across all classes. The numbers for other methods are reported from~\cite{topnet}. The results of the evaluation of our method against state-of-the-art methods are shown in Table~\ref{tab:chamferShapeNet}. We note that our result surpasses the previous works by a significant margin. Qualitative examples on ShapeNetCore data are given in Fig.~\ref{fig:shapenetres}. The distribution of distances given in Fig.~\ref{fig:chamferdist} implies that CC3D dataset presents advanced challenges for our autoencoder, where it performs at $1.26\times 10^{-3}$ average Chamfer distance, while it reaches $0.804\times10^{-3}$ on ShapeNetCore.
\begin{figure}[bt]
  \centering
 \includegraphics[width=0.7\linewidth]{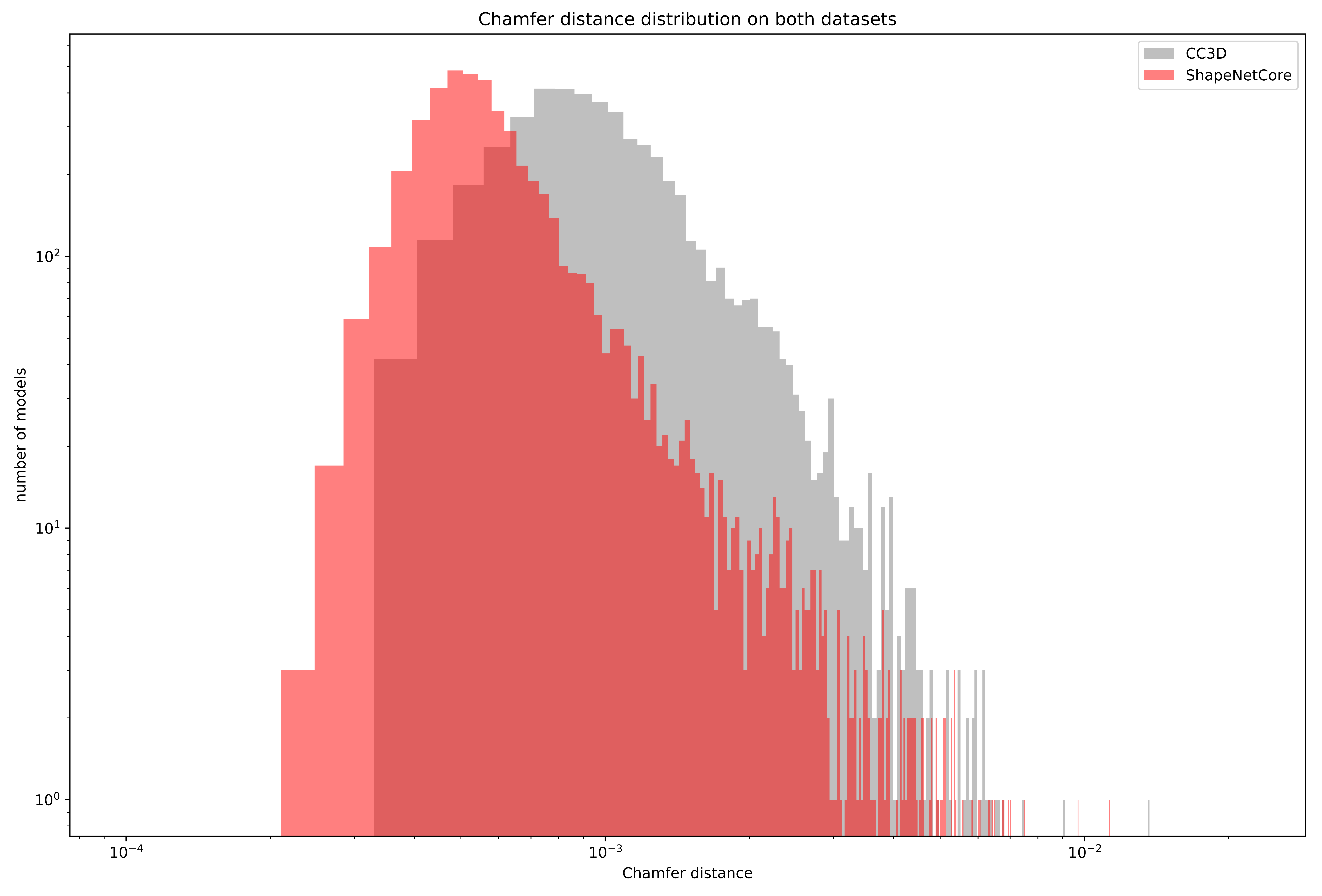}
  \vspace{-1em}
\caption{Chamfer distance distribution for our autoencoder. On test set of CC3D for point clouds of size $N=10$k, mean Chamfer distance is $1.26 \times 10^{-3}$ with standard deviation of $0.794 \times 10^{-3}$. ShapeNetCore test set with $N=2.5$k, it is $0.804 \times 10^{-3}$ with standard deviation $0.766 \times 10^{-3}$.}
\label{fig:chamferdist}
\vspace{-1em}
\end{figure}
\begin{figure}[t]
    \centering
    \includegraphics[width=0.9\linewidth]{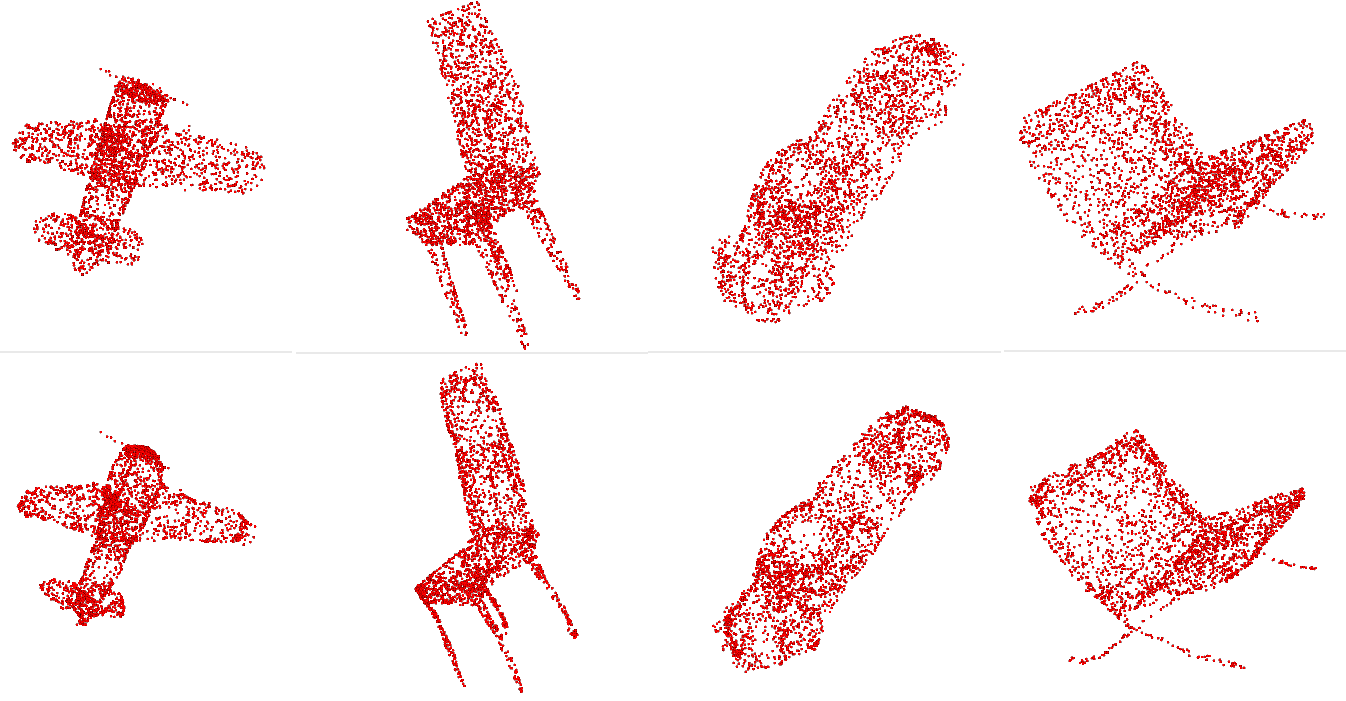}
    \vspace{-1em}
    \caption{Results of our autoencoder on ShapeNetCore data. The top row is the input 2.5k point clouds, the bottom is the reconstruction of our autoencoder.
    }
    \label{fig:shapenetres}
    \vspace{-1em}
\end{figure}


\section{CONCLUSIONS}
\label{sec:conclusions}
In this work, we proposed a Point-Voxel Deconvolution (\textit{PVDeConv}) block for a fast and efficient deconvolution on 3D point clouds. It was used in combination with a new dataset, CC3D, for autoencoding 3D Scans to their corresponding synthetic CAD models. The CC3D dataset offers pairs of CAD models and 3D scans, totaling to 50k+ objects. Our CC3D autoencoder on point clouds is memory and time efficient. Furthermore, it demonstrates superior results compared to existing methods on ShapeNet data. As future work, different types of losses will be investigated to improve the sharpness on edges, such as quadric~\cite{quadric}. 
Testing the variants of CC3D autoencoder with different configurations of stacked PVConv and PVDeConv layers will also be considered. Finally, we believe that the CC3D dataset itself could assist in real 3D scanned data analysis with deep learning methods.

\bibliography{main}
\bibliographystyle{IEEEbib}

\end{document}